\documentclass{article} 
\usepackage{amsmath}
\usepackage{color}
\usepackage{nips15submit_e,times}
\usepackage{hyperref}
\usepackage{url}
\usepackage{graphicx}
\usepackage{todonotes}
\usepackage{caption}
\usepackage{titlesec}
\usepackage{subcaption}

\usepackage[margin=0.2cm]{caption}
\title{Occam's Gates}

\author{
Jonathan Raiman \\
Massachusetts Institute of Technology\\
\texttt{jraiman@mit.edu} \\
\And
Szymon Sidor \\
Massachusetts Institute of Technology\\
\texttt{sidor@mit.edu}
}

\def\({\left(}
\def\){\right)}

\nipsfinalcopy 

\titlespacing*{\section}
{0pt}{10px}{0px}
\titlespacing*{\subsection}
{0pt}{5px}{0px}
\titlespacing*{\subsubsection}
{0pt}{5px}{0px}

\begin{document}


\maketitle

\begin{abstract}
We present a complimentary objective for training recurrent neural networks (RNN) with gating units that helps with regularization and interpretability of the trained model. Attention-based RNN models have shown success in many difficult sequence to sequence classification problems with long and short term dependencies, however these models are prone to overfitting. In this paper, we describe how to regularize these models through an L1 penalty on the activation of the gating units, and show that this technique reduces overfitting on a variety of tasks while also providing to us a human-interpretable visualization of the inputs used by the network. These tasks include sentiment analysis, paraphrase recognition, and question answering.
\end{abstract}

\section{Introduction}
Attention-based recurrent neural networks (RNN) have shown great success in a wide range of tasks such as computer vision
\cite{karpathy2014deep,vinyals2014show,mnih2014recurrent},
image generation
\cite{gregor2015draw,graves2013generating},
machine translation
\cite{bahdanau2014neural},
speech recognition
\cite{chorowski2014end},
or even as controllers for memory addressing and retrieval
\cite{graves2014neural,zaremba2015reinforcement}.

While there is debate as to how biologically plausible these cognition models are, they are desirable in their ability to allow introspection into the network's workings and understanding failures: in the case of image captioning \cite{vinyals2014show,karpathy2014deep}
and generation
\cite{gregor2015draw,graves2013generating},
or emotion detection \cite{zheng2014neural},
the system's focus matches up with human intuition.
The gates modulating the network's attention in these networks serve a dual purpose: first they allow control of the information flow, and second, and perhaps more crucially, the gates communicate problem structure by ensuring that specific groups of neurons activate or go dormant jointly.
For instance, in the case of prediction from a sequence of words, it is expected to find that certain words are predictive while others not; if this word sequence is projected using an embedding matrix into word vectors, then by the same logic all the dimensions of superfluous words' vectors should be wiped out entirely.

Intuitively, this Occam's Razor observation can be translated into considering that the activation of gating units should be as sparse as possible when not all the words or information units are necessary. The main focus of this paper is to show how to enforce sparsity on gating units by adding an unsupervised training objective: the sum of the activations of the gating units $g_i$ weighed by a hyper-parameter $\lambda_{\mathrm{sparse}}$ that controls the tradeoff between the original objective function $J$ and the sparsity criterion:
$$J^{*} = J + \lambda_{\mathrm{sparse}} \cdot \sum_{i} g_i.$$

In this work, we show that enforcing gate sparsity improves generalization in RNNs while also providing useful visualisations of the problem, and evaluate this approach on three different problems.

\section{Related Work}

The work we are presenting is closely related to two areas of Machine Learning research: RNN regularization and attention-based models.

\subsection{RNN Regularization}

RNN regularization has recently been shown to be achievable using {\em Dropout} \cite{srivastava2013improving} by regularizing a subset of the recurrent connections in deep RNNs \cite{zaremba2014recurrent,pham2013dropout}. Previously, it was shown that weight decay regularization only provided small improvement \cite{graves2013generating} and dropout noise was detrimental when applied to all connections due to the compounding of errors over time \cite{bayer2013fast}. In this work, we show that this problem can also be solved using a deterministic approach by penalizing gate activations from deep RNNs. As a result, RNNs can now benefit from multiple regularization techniques in varying architectures.

\subsection{Attention-Based Models}

In recent years, there has been a wealth of evidence that attention-based techniques can improve the performance of machine learning models.
Examples of this include work on capturing visual structure through a sequence of glimpses through images
\cite{gregor2015draw,tang2014learning,mnih2014recurrent,karpathy2014deep,vinyals2014show,zheng2014neural}, 
and networks that learn how to attend to and control a separate memory \cite{graves2014neural,WestonCB14,SukhbaatarSWF15}.

In certain cases the models are trained with supervision on the gates \cite{karpathy2014deep,WestonCB14}, however in many cases there is no supervised data for the attentional component. Several surrogate objectives have been suggested for learning where to focus, including setting a prior on observation spacing that makes a tradeoff between exploration and exploitation \cite{zheng2014neural}, using reinforcement learning \cite{zaremba2015reinforcement}
to optimize a visual tracking strategy \cite{mnih2014recurrent}, or leaving this part semi-supervised through the primary objective. Our work resembles the observation prior of \cite{zheng2014neural}, where we favor input gates being closed and penalize deviation with a penalty of our choosing. Similarly to the annealed Dropout from \cite{saon2015ibm}, we also consider a gradual increase in the sparsity penalty during training to encourage early exploration.
\section{Problem Statement}

A powerful family of models, often called Encoder-Decoders, have opened many new possibilities for sequence classification \cite{graves2013generating,graves2014neural,sutskever2014sequence}, including executing Python programs \cite{zaremba2014learning,chung2015gated}, drawing pictures \cite{gregor2015draw}, machine translation, or syntactic parsing \cite{vinyals2014grammar,DyerBLMS15}. The main problem we are trying to solve in this paper is improving generalization performance when performing these types of classical or structured prediction tasks using RNNs. In sections below we describe three different sequence classification problems used to evaluate our approach.

\subsection{Sentiment Analysis}

The central problem in sentiment analysis is correctly identifying and extracting the attitude or emotional tone of a speaker in the context of a particular topic or domain.

Here we consider predicting the sentiment expressed in the {\em Stanford Sentiment Treebank} (SST) \cite{rntn}, a collection of 11,855 sentences extracted from movie reviews. This dataset is made up of the sentiment annotations from 5 classes:  \{{\em terrible, bad, neutral, good, terrific}\}, for the 215,154 unique sub-phrases obtained after parsing each sentence using the Stanford Parser. In our work we do not make use of the parse trees, and instead treat each sub-phrase as a labeled sequence of words.

\subsection{Paraphrase Recognition}

In Paraphrase Recognition the problem is it to predict how semantically similar two phrases are from 0 to 1. This task can either be seen as regression or binary classification, and the goal is measured as the Pearson correlation with human annotations or recalling correct paraphrase pairs.

Here we focus on paraphrase detection on the SemEval 2014 shared task 1 dataset \cite{marelli2014semeval} which includes 9927 sentence pairs in a 4500/500/4927 train/dev/test split. Each sentence is annotated with a score $c \in [1,5]$, with 5 indicating the pair is a paraphrase, and 1 that the pair is unrelated. We additionally train using paraphrase pairs from the wikianswers paraphrase corpus \cite{parallex}.


\subsection{Question Answering}

Facebook AI Research recently proposed a set of 20 tasks designed to be ``prerequisites" for any system ``capable of conversing with human'' \cite{facebook2015babi}. The dataset for each task is a set of stories each  composed of {\em many facts}, with some marked as {\em relevant}, a {\em question} and the correct {\em answer}.

\begin{tabular}{p{.48\textwidth} p{.48\textwidth}}
\fbox{
\begin{minipage}{0.45\textwidth}
Daniel and Sandra journeyed to the office.\\
Then they went to the garden.\\
Sandra and John travelled to the kitchen.\\
After that they moved to the hallway.\\
Where is Daniel? {\bf \color{red} A: garden}
\end{minipage}
}
&
\fbox{
\begin{minipage}{0.45\textwidth}
The football fits in the suitcase.\\
The suitcase fits in the cupboard.\\
The box of chocolates is smaller than the football.\\
Will the box of chocolates fit in the suitcase?
{\bf \color{red} A:yes}
\end{minipage}
}
\end{tabular}

The tasks are synthetic and lack noisy nature of real-world natural processing, which makes them easy to solve with hand engineered systems, however the open question is how to create a model capable of solving these tasks without any manual feature engineering for particular problems.

\section{Approach}

In order to improve RNN performance over unseen data apply Occam's Razor over our training data by finding in each example a minimal set of useful inputs over time. To achieve this property we apply gates to the different observations of the input sequence to allow the network to keep or erase a timestep's input.
For instance, in a sentiment classification problem, gates would ideally fire only for emotionally loaded words, and stay dormant otherwise.

Because our approach relies on gates, we make the assumption that the vector input at each time-step is an inseparable  information unit, like a word, image, or fact. If this assumption holds, then when we force the network to reduce its gate usage by penalizing the sum of those activations, we will obtain a solution in a local optima where gates are less often active, which should  generalize better.

We formalise our approach by describing how we enforce sparsity on the gate activations for a variety of RNNs. Then we introduce the RNNs considered for the different tasks in this paper. Finally we explain the sparsity-enforcing objective function and our different annealing regimens during training.

\subsection{Gated LSTMs}
\label{gated_lstm}

In our work we make extensive use of Long-Short Term Memory networks  \cite{hochreiter1997long}, a popular RNN architecture specifically designed to capture long range dependencies and alleviate training difficulties \cite{pascanu2012difficulty}. 
Since their introduction in 1995, many variants have been proposed \cite{GreffSKSS15}, 
however for the purposes of this research we found that the {\em vanilla} version from \cite{GreffSKSS15} worked best.

\begin{table}[h!]
\caption{LSTM and Gated LSTM equations}
\renewcommand{\arraystretch}{1.2}
$$
\begin{array}{r|cll}
\mathrm{\bf description} & \mathrm{\bf symbol} & \mathrm{\bf LSTM} & \mathrm{\bf Gated\ LSTM} \\
\hline

\text{\color{red} Occam's\ gate} &
\color{red} g_{\mathrm{occam}} &
\text{absent} &
\color{red} f_{\mathrm{gate}}(\vec{x_t}, \vec{h}_{t-1}) \\

\text{\color{red} gated input} &
\color{red} \vec{x}_t' &
\text{absent} &
\color{red} \vec{x}_t \cdot  g_{\mathrm{occam}} \\

\text{block input}  &
\vec{z}_t &
\tanh\( \mathbf{W}_z \vec{x}_t + \mathbf{R}_z \vec{y}_{t-1} + \vec{b}_z\) &
\tanh\( \mathbf{W}_z  {\color{red}  \vec{x}_t' }  + \mathbf{R}_z \vec{y}_{t-1} + \vec{b}_z\)  \\

\text{input gate} &
\vec{i}_t &
\sigma\( \mathbf{W}_i \vec{x}_t + \mathbf{R}_i \vec{y}_{t-1} + \vec{b}_i\) &
\sigma\( \mathbf{W}_i   {\color{red}  \vec{x}_t' } + \mathbf{R}_i \vec{y}_{t-1} + \vec{b}_i\) \\

\text{forget gate}  &
\vec{f}_t &
\sigma\( \mathbf{W}_f \vec{x}_t + \mathbf{R}_f \vec{y}_{t-1} + \vec{b}_f\) &
\sigma\( \mathbf{W}_f  {\color{red}  \vec{x}_t' }  + \mathbf{R}_f \vec{y}_{t-1} + \vec{b}_f\) \\

\text{memory state} &
\vec{m}_t &
\vec{i}_t \odot \vec{z}_t + \vec{f}_t \odot \vec{m}_{t-1} &
\mathrm{identical} \\

\text{output gate} &
\vec{o}_t &
\sigma\( \mathbf{W}_o \vec{x}_t + \mathbf{R}_o \vec{y}_{t-1} + \vec{b}_o\) &
\sigma\( \mathbf{W}_o {\color{red}  \vec{x}_t' } + \mathbf{R}_o \vec{y}_{t-1} + \vec{b}_o\) \\

\text{hidden state} &
\vec{y}_t &
\vec{o}_t \odot \tanh\( \vec{c}_t \) &
\mathrm{identical} \\

\end{array}
$$
\label{lstm-eq}
\end{table}

While LSTMs are capable of selectively remembering or forget parts of their memory and input, they lack the ability to transform uniformly their input. We extend LSTMs to include an additional gate, $g_{\mathrm{occam}}$, that uniformly multiplies all the inputs simultaneously. In Table \ref{lstm-eq} we present equations for the Gated-LSTM, with the differences with the regular LSTM highlighted in red. We use the following denotations: $\sigma(\cdot)$ for the logistic sigmoid function, $\mathbf{W}_{i,z,f,o}$ and $\mathbf{R}_{i,z,f,o}$ for matrices, and $\vec{b}_{z,i,f,o}$ for vectors.

The gating function $f_{\mathrm{gate}}(\cdot)$ can take various forms. Two examples we consider are linear function of the input $\vec{x}_t$ and a second order gate capable of capturing higher-order interaction:
\begin{align*}
f_{\mathrm{linear}}(\vec{x}_t, \vec{h}_{t-1}) &= \sigma ( \vec{p}^T \cdot \vec{x}_t + \vec{q}^T \cdot \vec{h}_{t-1} + b)\\
f_{\mathrm{quad}}(\vec{x}_t, \vec{h}_{t-1}) &= \sigma ( \vec{h}^T \cdot \mathbf{W} \cdot  \vec{x}^T + \vec{p}^T \cdot \vec{x}_t + \vec{q}^T \cdot \vec{h}_{t-1} + b).
\end{align*}

Additionally, we consider Gated Stacked LSTMs, a variant of Stacked LSTMs \cite{graves2005framewise,graves2013generating,zaremba2014learning}, where the input the lowest LSTM is gated using the hidden state from the topmost LSTM of the previous timestep. The equation for this modification is as follows, with $l \in \{1, l_{\mathrm{max}}\}$, the LSTM level:
$$g_{\mathrm{occam}} = f_{\mathrm{gate}}(\vec{x}_{t}, \vec{h}_{( l_{\mathrm{max}}, t-1)}).$$

\subsection{Hierarchical Gated LSTMs}


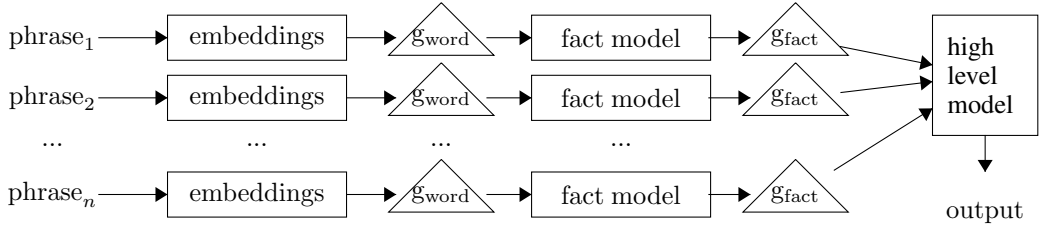
\begin{figure}[h!]
\begin{center}

\begin{tikzpicture}[scale=0.2]
\tikzstyle{every node}+=[inner sep=0pt]
\draw [black] (17.45,-1.9) rectangle (29.35,-4.9);
\draw (23.4,-3.4) node {$\mathrm{embeddings}$};
\draw [black] (17.45,-5.9) rectangle (29.35,-8.9);
\draw (23.4,-7.4) node {$\mathrm{embeddings}$};
\draw [black] (17.45,-12.333) rectangle (29.35,-15.333);
\draw (23.4,-13.83) node {$\mathrm{embeddings}$};
\draw [black] (41.65,-1.9) rectangle (53.55,-4.9);
\draw (47.6,-3.4) node {$\mathrm{fact\mbox{ }model}$};
\draw [black] (41.65,-5.9) rectangle (53.55,-8.9);
\draw (47.6,-7.4) node {$\mathrm{fact\mbox{ }model}$};
\draw [black] (41.65,-12.333) rectangle (53.55,-15.333);
\draw (47.6,-13.83) node {$\mathrm{fact\mbox{ }model}$};
\draw [black] (68.3,-1.9) rectangle (75.3,-9.9);
\draw [text width=1cm] (71.8,-5.9) node {high level model};
\draw (9.8,-3.4) node {$\mathrm{phrase}_1$};
\draw (9.8,-7.4) node {$\mathrm{phrase}_2$};
\draw (9.8,-13.83) node {$\mathrm{phrase}_n$};
\draw (9.8,-10.5) node {$\mathrm{...}$};
\draw (35.65,-10.5) node {$\mathrm{...}$};
\draw (47.6,-10.5) node {$\mathrm{...}$};
\draw [black] (59.2,-1.07) -- (62.7,-4.57) -- (55.7,-4.57) -- (59.2,-1.07);
\draw (59.2,-3.4) node {$\mathrm{g}_{\mathrm{fact}}$};
\draw [black] (59.2,-5.07) -- (62.7,-8.57) -- (55.7,-8.57) -- (59.2,-5.07);
\draw (59.2,-7.4) node {$\mathrm{g}_{\mathrm{fact}}$};
\draw [black] (59.2,-11.5) -- (62.7,-15) -- (55.7,-15) -- (59.2,-11.5);
\draw (59.2,-13.83) node {$\mathrm{g}_{\mathrm{fact}}$};
\draw [black] (35.65,-1.07) -- (39.15,-4.57) -- (32.15,-4.57) -- (35.65,-1.07);
\draw (35.65,-3.4) node {$\mathrm{g}_{\mathrm{word}}$};
\draw [black] (35.65,-5.07) -- (39.15,-8.57) -- (32.15,-8.57) -- (35.65,-5.07);
\draw (35.65,-7.4) node {$\mathrm{g}_{\mathrm{word}}$};
\draw [black] (35.65,-11.5) -- (39.15,-15) -- (32.15,-15) -- (35.65,-11.5);
\draw (35.65,-13.83) node {$\mathrm{g}_{\mathrm{word}}$};
\draw (23.4,-10.5) node {$\mathrm{...}$};
\draw (71.8,-15) node {$\mathrm{output}$};
\draw [black] (29.35,-3.4) -- (32.65,-3.4);
\fill [black] (32.65,-3.4) -- (31.85,-2.9) -- (31.85,-3.9);
\draw [black] (38.65,-3.4) -- (41.65,-3.4);
\fill [black] (41.65,-3.4) -- (40.85,-2.9) -- (40.85,-3.9);
\draw [black] (29.35,-7.4) -- (32.65,-7.4);
\fill [black] (32.65,-7.4) -- (31.85,-6.9) -- (31.85,-7.9);
\draw [black] (38.65,-7.4) -- (41.65,-7.4);
\fill [black] (41.65,-7.4) -- (40.85,-6.9) -- (40.85,-7.9);
\draw [black] (29.35,-13.83) -- (32.65,-13.83);
\fill [black] (32.65,-13.83) -- (31.85,-13.33) -- (31.85,-14.33);
\draw [black] (38.65,-13.83) -- (41.65,-13.83);
\fill [black] (41.65,-13.83) -- (40.85,-13.33) -- (40.85,-14.33);
\draw [black] (12.85,-3.4) -- (17.45,-3.4);
\fill [black] (17.45,-3.4) -- (16.65,-2.9) -- (16.65,-3.9);
\draw [black] (12.85,-7.4) -- (17.45,-7.4);
\fill [black] (17.45,-7.4) -- (16.65,-6.9) -- (16.65,-7.9);
\draw [black] (12.85,-13.83) -- (17.45,-13.83);
\fill [black] (17.45,-13.83) -- (16.65,-13.33) -- (16.65,-14.33);
\draw [black] (53.4,-3.4) -- (56.2,-3.4);
\fill [black] (56.2,-3.4) -- (55.4,-2.9) -- (55.4,-3.9);
\draw [black] (53.4,-7.4) -- (56.2,-7.4);
\fill [black] (56.2,-7.4) -- (55.4,-6.9) -- (55.4,-7.9);
\draw [black] (53.4,-13.83) -- (56.2,-13.83);
\fill [black] (56.2,-13.83) -- (55.4,-13.33) -- (55.4,-14.33);
\draw [black] (71.8,-9.87) -- (71.8,-12.33);
\fill [black] (71.8,-12.33) -- (72.3,-11.53) -- (71.3,-11.53);
\draw [black] (62.14,-3.98) -- (68.3,-5.21);
\fill [black] (68.3,-5.21) -- (67.61,-4.56) -- (67.42,-5.54);
\draw [black] (62.18,-7.05) -- (68.3,-6.32);
\fill [black] (68.3,-6.32) -- (67.45,-5.91) -- (67.56,-6.91);
\draw [black] (61.74,-12.23) -- (68.3,-8.1);
\fill [black] (68.3,-8.1) -- (67.36,-8.11) -- (67.89,-8.95);
\end{tikzpicture}

\end{center}
\caption{Architecture for Hierarchical Gated-LSTMs}
\label{fig:hg_lstm}
\end{figure}

In this section we introduce Hierarchical Gated LSTM (HG-LSTM), a gated attention model that uses Gated LSTMs as a central building block. In the previous section we introduced Gated LSTMs that are able to selectively ignore or include the entire input at a timestep, however for many tasks where the information presented can be subdivided into larger chunks such as sentences, paragraphs, or episodes, a similar gating procedure could be applied to these higher levels of abstraction.
For example to find the answer to question about a story in the bAbI dataset, such a model would benefit from being selective about which words and facts to listen to.

HG-LSTM consists of two submodels: a {\em Fact model} and {\em High-Level model (HL model)}, which are both Gated LSTMs. Figure \ref{fig:hg_lstm} presents the architecture. Every word in a fact sequence is projected using an embedding matrix and processed by the {\em Fact model}. The final hidden state of the {\em Fact model} for each fact is then passed to the {\em HL model} as an input vector. We consider the final hidden state of the {\em HL model} after reading each fact representation to be a the high-level representation for the entire sequence of facts. The hierarchy of the submodels explicitly leverages the problem structure, and  allows fine grain attention control at two levels of abstraction.

\subsection{Sparsity Penalty}
The original training objective $J$ is augmented with the sparsity penalty and the resulting objective is optimized through gradient descent. The penalty is constructed by summing the activations of the gates presented in \ref{gated_lstm}, and weighing them by a parameter $\lambda_{\mathrm{sparse}}$ chosen through hyperparameter search:
$$
J^{*} = J + \lambda_{\mathrm{sparse}} \cdot \sum_{i=1}^n g_{\mathrm{occam}, i}.
$$

\subsection{Training Regimens}
\label{annealing}

Our approach's ultimate goal is to preserve network expressivity while making it robust against changes in the input. However, forcing sparsity too soon can do more harm than good: a greedy and locally optimal solution is forcing all gates to be closed. To prevent this from happening we encourage early exploration by progressively increasing the sparsity penalty, $\lambda_{\mathrm{sparse}}$. We  investigated 2 different annealing regimens: a linear and a quadratic increase up to $\lambda_{\mathrm{max}}$ at training epoch $T_{\mathrm{max}}$, as shown below with $e$ the training epoch:
$$
\lambda_{\mathrm{sparse}}(e) = \begin{cases}
\lambda_{max} & \text{flat regimen}\\
\min \{ (e / T_{\mathrm{max}}) \cdot \lambda_{max} , \lambda_{max} \} & \text{linear regimen}\\
\min \{ (e / T_{\mathrm{max}})^2 \cdot \lambda_{max} , \lambda_{max} \} & \text{quadratic regimen}\\
\end{cases}$$
\section{Experiments}
\ifnipsfinal
The code needed to run the experiments in this paper are available online at
\url{https://www.github.com/JonathanRaiman/Dali} \footnote{The project is currently under heavy development, do not hesitate to ask the authors for help!}.
\fi


\subsection{Sentiment Analysis}

For this problem our model is a Gated LSTM that reads each sequence of words sequentially, and uses the last hidden vector as input to a softmax linear classifier, and our target is to minimize the Kullback-Leibler divergence with the correct label along with the sparsity penalty.

We project each word using an embedding matrix into a 100 dimensional vector, and keep only the words that appear at least twice in our training data, with the remaining words replaced with a special unknown word, {\tt <UNK>}. We train 3 different models with hidden sizes 25, 50, 150, and apply Dropout \cite{srivastava2013improving,zaremba2014recurrent} with probability $p=0.3$ to the non recurrent connections of the LSTM.
All models are trained using Adadelta \cite{zeiler2012adadelta} with $\rho=0.95$, and we perform early stopping when the accuracy stops increasing on the validation set.
\subsection{Paraphrase Detection}

For paraphrase prediction we also employ Gated LSTMs with the final Softmax layer removed. Each sentence in a pair is fed to a separate LSTM and our objective is to minimize the squared difference between the true similarity $t$ of the sentences and the dot product of the two LSTMs' final hidden states $\vec{h}_1, \vec{h}_2$:
\begin{center}$J = \mathrm{min}\left\{
\(
\frac{
\vec{h}_{1}^T \dot \vec{h}_{2}
}{
|\vec{h}_{1}| \dot |\vec{h}_{2}|
} - t
\)^2
\right\} + \lambda_{\mathrm{sparse}} \cdot \(\sum_{i = 1}^n g_{1, i} + \sum_{i = 1}^n g_{2, i} \)$\end{center}

 instead of a softmax linear classifier, we instead use the last hidden state of the LSTM.

\subsection{Facebook's bAbI dataset}

For this problem we use an HG-LSTM to compute the high level representation of each story. The HG-LSTM takes a {\em question}, followed by the {\em sequence of facts}, and the final hidden state of the HG-LSTM is fed as input to an LSTM decoder that produces the answer sequentially and ends its prediction with an {\tt <EOS>} symbol \cite{graves2013generating,vinyals2014grammar}.

We use separate a Gated-LSTM for question and facts when creating representations for the {\em High-Level model} in the HG-LSTM. To make the question influence the High Level's input gates we average the embeddings of the words in the question and concatenate this with the fact representation and the current hidden state of the {\em High Level model}.

Our error function is the sum of three separate objectives:
\begin{align*}
E_{\mathrm{prediction}} =& \sum_{w \in Y} \sum_{\bar{w} \neq w} \mathrm{max}(\gamma - s(w) + s(\bar{w}), 0)\\
E_{\mathrm{fact}} =& \sum_{i \in F} \mathrm{log}(g_i) + \mu_{\mathrm{unsupporting}} \sum_{i \notin F} \mathrm{log}(1 - g_i)\\
E_{\mathrm{word}} =& \sum_{f \in F} \sum_{w \in f} |g_w|
\end{align*}

Prediction error $E_{\mathrm{prediction}}$ defined as margin loss on every word of the output,where $Y$ is a target sequence of words, $s(w)$ is a score a particular word and $\gamma$ is margin. We found that it significantly decreases training time compared to cross entropy error while achieving similar results.\\
For fact selection error $E_{\mathrm{fact}}$ a set of supporting facts $S$ is known, therefore rather than using sparsity penalty, we used cross entropy error between expected ($1$ for $f \in S$ and $0$ otherwise) and actual gate activation. $F$ is set of fact indexes, $g_i$ is activation of gate for fact $i$. The $\mu_{\mathrm{unsupporting}}$ coefficient was introduced because authors reasoned that false negatives are potentially more harmful than false negatives for network learning process.\\
Finally $E_{\mathrm{word}}$ is a L1 sparsity penalty for all the word gates in fact model. Symbol $g_w$ denotes gate activation for a particular word in a particular fact. \\
We combine the errors into a single objective:
\begin{align*}
E = E_{\mathrm{prediction}} + \lambda_{\mathrm{fact}}E_{\mathrm{fact}} + \lambda_{\mathrm{word}} E_{\mathrm{word}}
\end{align*}

Our precise parameters for the experiment were as follows: all word embeddings have 50 dimensions, we used Dropout with $p= 0.5$ in the {\em High Level model} and $p=0.3$ for {\em Question} and {\em Fact models}. The {\em Fact model} has a hidden size of 30, while the {\em High level model} is a Gated Stacked-LSTM with 6 layers and a hidden size of 20. All the gates used are second order, $f_{\mathrm{quad}}(\cdot)$.

We use the first 1000 examples for training as suggested in \cite{facebook2015babi}, and reserve 20\% for validation. Our model is trained using AdaDelta \cite{zeiler2012adadelta}, with $\rho = 0.95$, and a minibatch size of 50. We perform early stopping when the validation score stops increasing.

\section{Results}
\subsection{Effects on performance}
\begin{figure}[h!]
\centering
    \begin{minipage}{0.49\textwidth}
      	\centering
        \includegraphics[width=\textwidth]{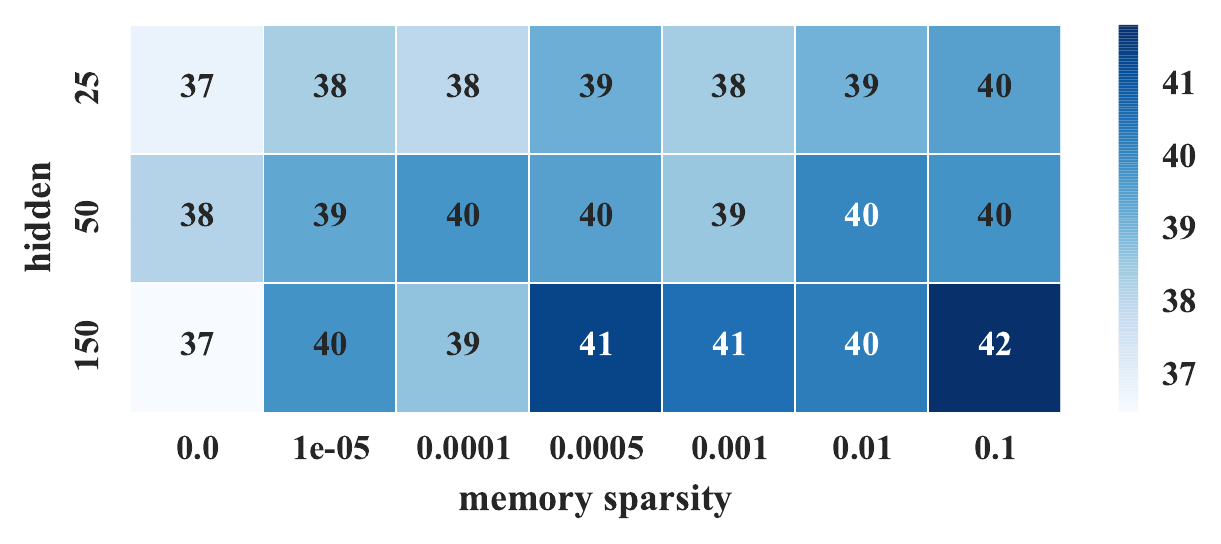}
      \captionof{figure}{SST Root Accuracy with varying LSTM hidden size and sparsity penalty $\lambda$}
      \label{fig:sentimentHidden}
    \end{minipage}
    \begin{minipage}{0.49\textwidth}
      \centering
        \includegraphics[width=\textwidth]{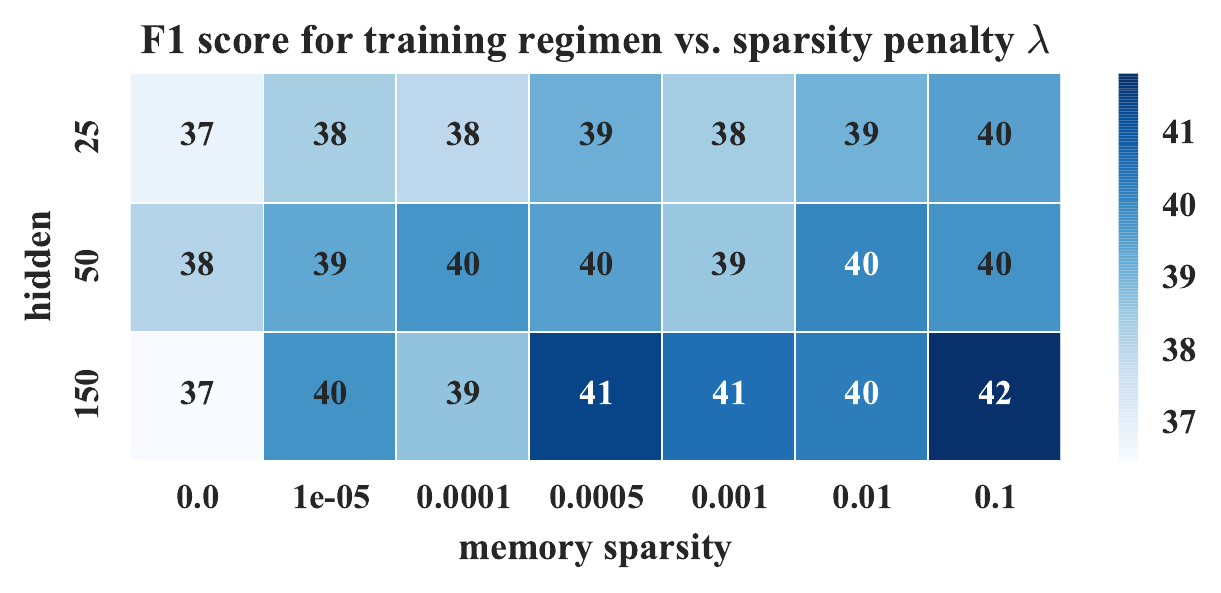}
      \captionof{figure}{Effect of sparsity regimen and sparsity penalty $\lambda$ on SST Root Accuracy.}
      \label{fig:sentimentRegimen}
    \end{minipage}
\end{figure}

\begin{figure}[h!]
    \centering
    \begin{minipage}{.49\textwidth}
      \centering
        \includegraphics[width=\textwidth]{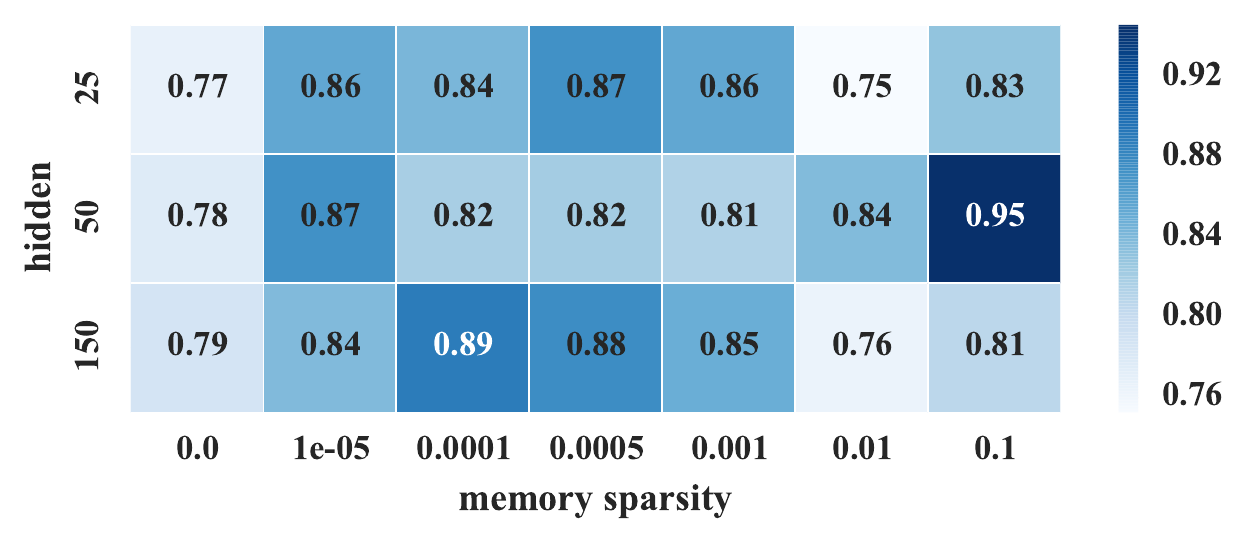}
      \captionof{figure}{Paraphrase accuracy with varying LSTM hidden size and sparsity penalty $\lambda$}
      \label{fig:paraphraseHidden}
    \end{minipage}
    \begin{minipage}{.49\textwidth}
      \centering
        \includegraphics[width=\textwidth]{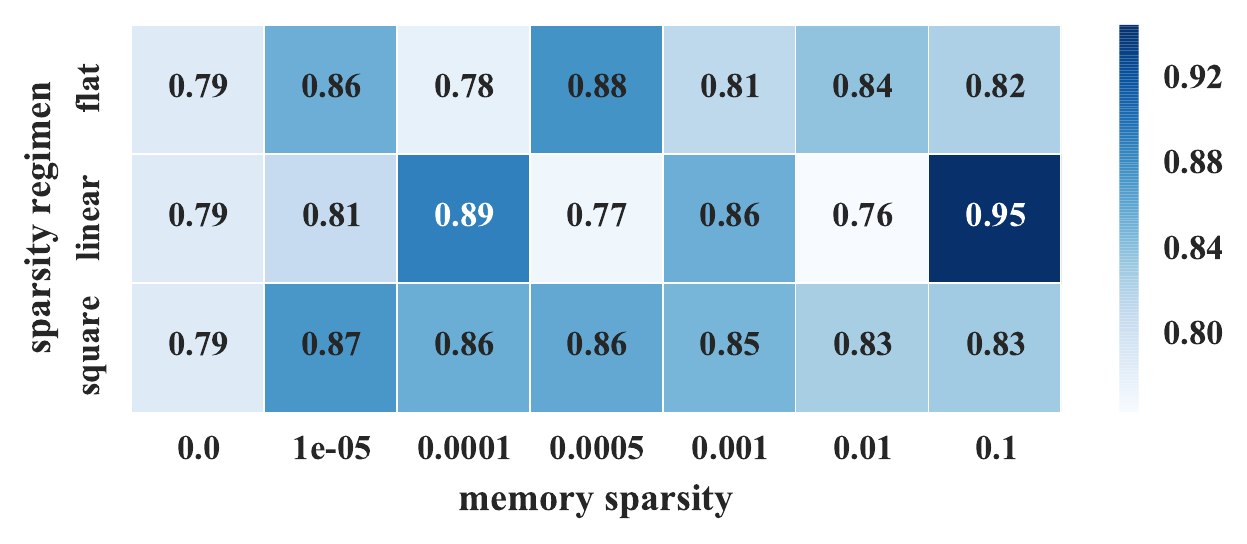}
      \captionof{figure}{Effect of sparsity regimen and penalty $\lambda$ on Paraphrase prediction.}
      \label{fig:paraphraseRegimen}
    \end{minipage}
\end{figure}

\begin{figure}[h!]
    \centering
    \begin{minipage}{.49\textwidth}
      \centering
        \includegraphics[width=\textwidth]{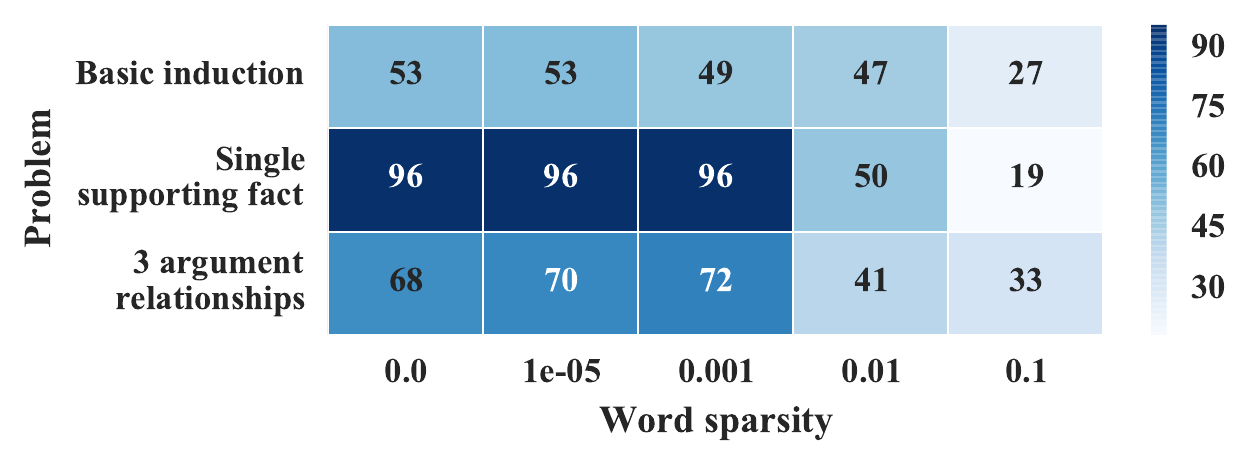}
      \captionof{figure}{Accuracy for three bAbI tasks with varying $\lambda_{\mathrm{word}}$}
      \label{fig:babi1}
    \end{minipage}
    \begin{minipage}{.49\textwidth}
      \centering
        \includegraphics[width=\textwidth]{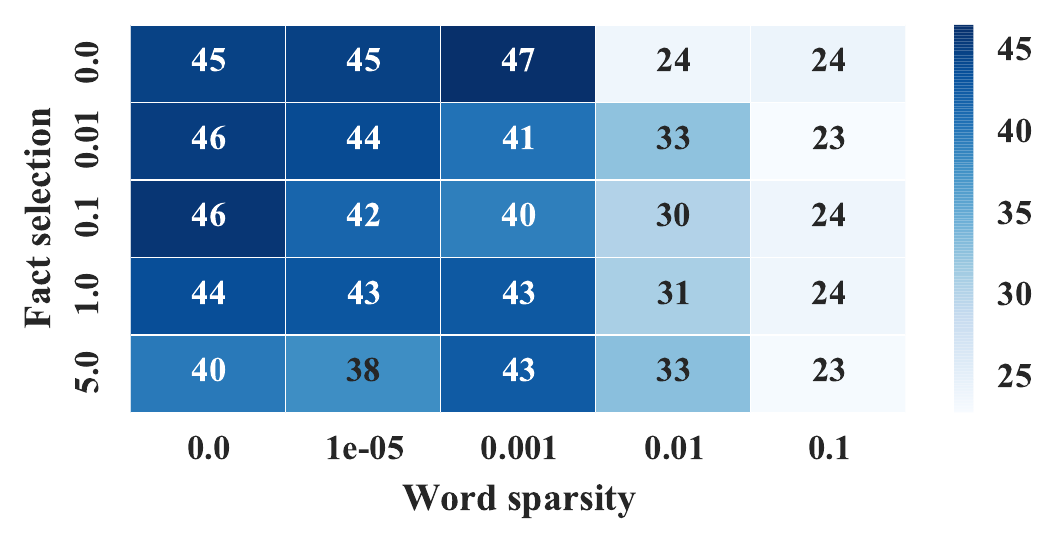}
      \captionof{figure}{Effect of $\lambda_{\mathrm{word}}$ and $\lambda_{\mathrm{word}}$ on Basic Induction task accuracy}
      \label{fig:babi2}
    \end{minipage}
\end{figure}

Occam's gates improve generalization on sentiment analysis (fig. \ref{fig:sentimentHidden}), paraphrase prediction (fig. \ref{fig:paraphraseHidden}), and for the majority of bAbI question answering problems (fig. \ref{fig:babi1}, Table \ref{tabula-rasa}). This effect is especially visible as model size increases (fig. \ref{fig:sentimentHidden}, fig. \ref{fig:paraphraseHidden}). We find that without a sparsity penalty increasing model size has smaller effect, however using sparsity we manage to achieve 5\% improvement on sentiment analysis and 18\% on paraphrase prediction recall. Additionally for {\em three arg. relations} bAbI problem it increases the accuracy by 14\%.
We observe greater improvements on this task than the other two; notably, this task has longer sentences, and thus word gating is more present.

Moreover, the sparsity annealing methods described in section \ref{annealing} show improvements over a static objective function (fig. \ref{fig:sentimentRegimen}, fig. \ref{fig:paraphraseRegimen}). In particular, the linear regimen improves the result by 1\% for sentiment analysis, and by 7\% for recall on paraphrase prediction.

Finally, we observed that the HG-LSTM model significantly improves performance over the LSTM baseline from \cite{facebook2015babi}. As visible in table \ref{tabula-rasa}, this model improves scores on 17 out of 20 problems. Moreover, HG-LSTMs with no penalties, $\lambda_{word} = \lambda_{fact} = 0$, yields worse results than those with penalties for the majority of the problems (17 out of 20 tasks).
Our best results are achieved by using mixture of both fact detection penalisation and word sparsity (7 out of 20 task). The HG-LSTM performs worse than Memory Networks (MemNN), however our model appears to be less computationally costly since we do not require branch and bound search to select supporting facts.

\renewcommand{\arraystretch}{0.75}

\begin{table}[t]
\caption{Comparison of test accuracy on bAbI dataset from \cite{facebook2015babi} with different models. Models are (left to right): LSTM baseline from \cite{facebook2015babi}, followed HG-LSTM with: no penalty, word sparsity penalty only, fact selection penalty only and both. The last column is MemNN.}
\label{tabula-rasa}
\begin{center}
\begin{tabular}{lp{1cm}p{1.4cm}p{1cm}p{1cm}p{1cm}|p{1.5cm}}
\bf task & \bf LSTM & \bf No penalty & \bf Word Penalty & \bf Fact Penalty & \bf Fact, word  & \bf MemNN\\
single supporting fact  & 50          & 81            &  45      & \bf 100 &     99          & 100 \\
two supporting facts    & 20          & 32            &  19      & 30      & \bf 32          & 100 \\
three supporting facts  & 20          & 19            &  20      & 16      & \bf 20          & 100 \\
two arg relations       & 61          & 76            &  65      & 76      & \bf 77          & 100 \\
three arg relations     & \bf 70      & 51            &  66      & 40      & 31              & 98  \\
yes-no questions        & 48          & 48            & \bf 51   & 50      & 50              & 100 \\
counting                & 49          & \bf 76        &  65      & 69      & 70              & 85  \\
lists sets              & 45          & \bf 78        &  66      & 76      & 73              & 91  \\
simple negation         & 64          & 67            &  65      & \bf 70  &  69             & 100 \\
indefinite knowledge    & 44          & 45            &  \bf 47  & 40      &  44             & 98  \\
basic-coreference       & 72          & 87            &  50      & 88      & \bf 89          & 100 \\
conjunction             & 74          & 75            &  66      & 99      & \bf 99          & 100 \\
compound-coreference    & \bf 94      & 73            &  93      & 91      & 86              & 100 \\
time reasoning          & 27          & \bf 27        &  19      & 18      & 18              & 99  \\
basic deduction         & 21          & 39            & \bf  50  & 24      & 50              & 100 \\
basic induction         & 23          & 44            &  42      & \bf 47  &     40          & 100 \\
positional reasoning    & 51          & 52            &  52      & 52      & \bf 58          & 65  \\
size reasoning          & 52          & 54            &  \bf 90  & 89      & 50              & 95  \\
path finding            & \bf 8       & 8             &  8       & 8       & 8               & 36  \\
agents motivations      & 91          & 95            &  63       & 66      & \bf 96          & 100 \\
\end{tabular}
\end{center}
\label{tab:babi_results}
\end{table}


\subsection{Interpretability}
Ability to interpret the calculation carried out by Machine Learning models is crucial for advancing research. Especially for Neural Network models there are no well established methods for understanding its capabilities, although attempts have been made, e.g Hinton Diagrams \cite{bremner1994hinton}. We claim that Occam's Razors provide some insights into the way network operates on it's hidden state.

\subsubsection{Error analysis}

Diagnosing and identifying the root cause of errors during model design is critical for finding with new research directions and making improvements. We believe using {\em Occam's gates} can help researchers gain insight into their network's workings. To support this claim let us consider an example from bAbi dataset where gates provide a visual indication of progress.

\begin{figure}[h!]
    \centering
    \begin{minipage}{.32\textwidth}
      \centering
        \includegraphics[width=.95\textwidth]{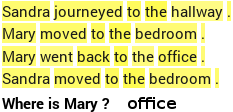}
    \end{minipage}
    \begin{minipage}{.32\textwidth}
      \centering
        \includegraphics[width=.95\textwidth]{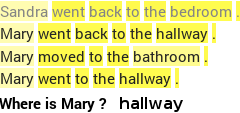}
    \end{minipage}
    \begin{minipage}{.32\textwidth}
      \centering
        \includegraphics[width=.95\textwidth]{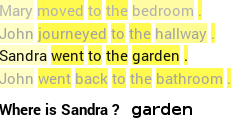}
    \end{minipage}
    \caption{Example story from the single supporting fact bAbI. Activation of word gates is shown with yellow highlighter. Text opacity reflects the activation of the fact gate for the sentence. The images were taken when validation accuracy was 20\%, 60\% and 100\% (left to right). }
    \label{fig:babi_errors}
\end{figure}

In Figure \ref{fig:babi_errors} we notice that the model upon reaching a validation accuracy of  20\% is not yet capable of distinguishing important information from noise.
At 60\% accuracy it can now highlights the relevant facts, but the gates on words are not yet compelling. At 100\% accuracy fact and word gates work in unison: the network activates for fact with the relevant person and words that contain {\em location} information. We hypothesize that LSTMs without gates can pick out the correct person and place, but {\em Occam's gates} help them ignore facts about persons irrelevant to the question.

\subsubsection{Relevancy detection}
\begin{figure}[h!]
    \centering
    \begin{minipage}{.49\textwidth}
      \centering
        \includegraphics[width=.9\textwidth]{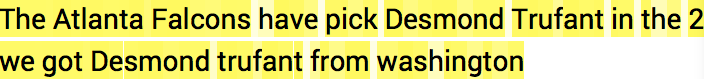}
      \captionof{figure}{Char Gated LSTM, gate action shown with yellow highlighter. Model discovers tokenisation.}
      \label{fig:tokenizing}
    \end{minipage}
    \begin{minipage}{.49\textwidth}
      \centering
        \includegraphics[width=.9\textwidth]{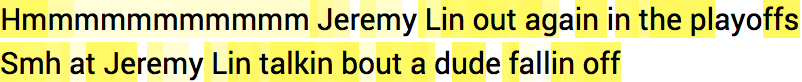}
      \captionof{figure}{Char Gated LSTM, gate action shown with yellow highlighter. Model focuses on upper case characters and ignores repeats.}
      \label{fig:hmm}
    \end{minipage}
\end{figure}

We argue that {\em Occam's gates} allow one to judge which pieces of information are relevant to a  problem. To illustrate this claim we show two examples, both of which emerged when training the system on a paraphrase detection problem with a Character model Gated LSTM (Char Gated LSTM). Figure \ref{fig:tokenizing} supports the belief that the model makes use word boundaries, and figure \ref{fig:hmm} suggests that the network can ignore repetititve or superfluous characters.

\section{Conclusion}

In this paper, we investigated the use of a complimentary objective function that forces attention-based RNNs to be selective about their inputs.
We showed on three different tasks that our approach improves generalization and interpretability of the trained models with respect their counterparts that do not use sparsity penalties.
Additionally, to encourage early exploration and preserve sparsity, we designed an annealing objective function that provides benefits over a standard one.

Finally, we introduced Hierarchical-Gated LSTM, a new model that performs significantly better than regular Stacked LSTMs; this network combines attentional and hierarchical components, and reasons at several levels of abstraction. Future work includes investigation of this model family, which shows promise towards advancing the state of the art.

\subsubsection*{References}

\bibliographystyle{unsrt}

\begingroup
\renewcommand{\section}[2]{}
{\small
\bibliography{references}
}
\endgroup


\end{document}